\documentclass[11pt]{article}

\usepackage[final]{acl}

\usepackage{times}
\usepackage{latexsym}

\usepackage[T1]{fontenc}

\usepackage[utf8]{inputenc}

\usepackage{microtype}

\usepackage{inconsolata}

\usepackage{graphicx}
\usepackage{amsfonts}
\usepackage{amsmath}
\usepackage{amsthm}
\usepackage{bm}
\usepackage{booktabs}
\usepackage{multirow}
\usepackage{adjustbox}
\usepackage{subcaption}
\usepackage{array}
\usepackage{xcolor}
\newtheorem{definition}{Definition}
\newcommand{\myparagraph}[1]{\vspace{0.5mm} \noindent\textbf{#1}}
\newcommand{\opt}[1]{%
  \colorbox{yellow!25}{#1}%
}
\title{HiTeC: Hierarchical Contrastive Learning on Text-Attributed Hypergraph with Semantic-Aware Augmentation}

\author{
Mengting Pan\textsuperscript{1},
Fan Li\textsuperscript{1},
Chen Chen\textsuperscript{2},
Xiaoyang Wang\textsuperscript{1},
Wenjie Zhang\textsuperscript{1} \\
\textsuperscript{1}The University of New South Wales \\
\textsuperscript{2}University of Wollongong \\
\texttt{\{mengting.pan,fan.li8,xiaoyang.wang1,wenjie.zhang\}@unsw.edu.au} \\
\texttt{chenc@uow.edu.au}
}

\begin{document}
\maketitle
\begin{abstract}
Contrastive learning (CL) has become a dominant paradigm for self-supervised hypergraph learning, enabling effective training without costly labels. However, node entities in real-world hypergraphs are often associated with rich textual information, which has been largely ignored in prior works. Directly applying existing CL-based methods to such text-attributed hypergraphs (TAHGs) leads to three key limitations: (1) The common use of graph-agnostic text encoders fails to capture the correlations between textual semantics and hypergraph topology, resulting in less expressive representations. (2) Their reliance on random data augmentations introduces noise and weakens the contrastive signals. (3) The primary focus on node- and hyperedge-level contrastive signals limits the ability to capture long-range dependencies, which is essential for effective representation learning. To address these challenges, we introduce HiTeC, a two-stage hierarchical contrastive learning framework for effective self-supervised learning on TAHGs. In the first stage, we pre-train the text encoder with a structure-aware contrastive objective to overcome the graph-agnostic nature of conventional methods. In the second stage, we begin by introducing semantic-aware augmentations, including structure-contextualized text augmentation and semantic-aware hyperedge dropping, to facilitate informative view generation. Subsequently, we propose a multi-scale contrastive loss with an $s$-walk-based subgraph-level objective to capture long-range dependencies. Extensive experiments on six real-world datasets validate the effectiveness of our proposed method.
\end{abstract}

\section{Introduction}
Hypergraphs extend traditional graphs by allowing each hyperedge to connect multiple nodes, offering a natural way to model multi-way relationships.
Such higher-order structures are prevalent in complex real-world systems, e.g., cortical co-activation in brains~\cite{yu2011higher}, user-item interactions in e-commerce platforms~\cite{recommender2022}, and multi-author collaborations in academic networks~\cite{collaboration2023}.
In these scenarios, nodes are often associated with rich textual attributes (e.g., product reviews in user–item networks and paper abstracts in co-authorship networks), which provide essential semantic information for downstream applications.
While Hypergraph Neural Networks (HNNs) have emerged as a powerful paradigm for hypergraph learning~\cite{AllSetTransformer,kim2024survey}, effectively capturing high-order structural dependencies, they primarily rely on numerical features and task-specific labels. 
This reliance limits their effectiveness in real-world applications, where labeled data is scarce, and node attributes are predominantly textual.
This motivates the development of self-supervised learning (SSL) methods tailored to text-attributed hypergraphs (TAHGs)~\cite{hyperbert2024,tahg}.

Recent advances in hypergraph self-supervised learning (HSSL) primarily focus on the contrastive paradigm, which maximizes agreement between two augmented views derived from the original hypergraph. 
HyperGCL~\cite{hypergcl2022} pioneers this direction with a learnable augmentation strategy. 
TriCL~\cite{tricl2023} introduces tri-directional contrastive objectives over node-, hyperedge-, and membership-level relations in hypergraphs. 
SE-HSSL~\cite{se-hssl2024} further redesigns a hierarchical membership-level objective, achieving state-of-the-art (SOTA) performance.
Beyond contrastive methods, generative approaches have also gained initial attention. HypeBoy~\cite{hypeboy2024} proposes a hyperedge-filling pretext task to learn expressive hypergraph representations.

Despite recent progress, current HSSL approaches still face significant limitations on hypergraphs with textual attributes: 
(1) They typically use shallow encoders (e.g., Bag-of-Words~\cite{bag2010} and Skip-Gram~\cite{skip2013}) or pre-trained language models (e.g., BERT~\cite{bert2019}) to convert raw text into numerical features while ignoring hypergraph structure.
This graph-agnostic design overlooks the correlation between text and higher-order topology, leading to suboptimal representations~\cite{giant2022}.
(2) Current hypergraph contrastive learning (HCL) methods often apply random augmentations to features and structure, e.g., random masking of text or hyperlinks. 
These perturbations may distort textual meaning and disrupt the alignment between structure and semantic content, introducing noise into the contrastive objective~\cite{gaugllm2024}.
(3) Existing HCL methods mainly emphasize node- and hyperedge-level signals, but struggle to capture long-range dependencies and richer topological contexts that are crucial for expressive representation learning~\cite{hash-code2024}.

HyperBERT~\cite{hyperbert2024} takes an early attempt at SSL on TAHGs by jointly training text and hypergraph encoders through contrastive learning. However, its contrastive objective mainly relies on 1-hop hyperedge relations, limiting its ability to capture multi-scale dependencies that have proven beneficial for HSSL~\cite{tricl2023,se-hssl2024}.
Meanwhile, recent text-attributed graph studies~\cite{giant2022,gaugllm2024,hash-code2024} explore joint text--structure modeling, but their extension to TAHGs remains challenging due to insufficient modeling of higher-order relations inherent to hypergraphs.

To address these issues, we propose HiTeC, a hierarchical contrastive learning framework for effective SSL on TAHGs. 
HiTeC adopts a two-stage framework that first injects higher-order topology into text representations and then performs semantic-aware hypergraph contrastive learning.
In the first stage, we pre-train a text encoder with a structure-aware contrastive objective to align textual semantics with higher-order topology, producing topology-enhanced text representations. In the second stage, we pre-train a hypergraph encoder with semantic-aware augmentations and hierarchical contrastive objectives. Specifically, we introduce structure-contextualized text augmentation and semantic-aware hyperedge dropping to alleviate perturbation noise. Based on the augmented views, we further formulate node-, hyperedge-, and subgraph-level contrastive objectives, where an \(s\)-walk-based subgraph signal captures long-range dependencies and richer structural contexts.
Our main contributions are as follows:
\begin{itemize}
    \item We propose HiTeC, an effective two-stage contrastive learning framework for TAHGs that addresses key limitations of existing HSSL methods in modeling rich textual attributes and complex hypergraph structure.
   \item We introduce a structure-aware contrastive objective for text encoder pretraining to integrate textual content with hypergraph structure.
   \item We enhance the hypergraph encoder with two semantic-aware augmentation strategies and hierarchical contrastive objectives, where structure-contextualized text augmentation and semantic-aware hyperedge dropping are used to reduce perturbation noise, and an \(s\)-walk-based subgraph-level contrastive signal captures long-range structural dependencies.
    \item Extensive experiments on six real-world datasets demonstrate the effectiveness and superiority of HiTeC over strong baselines across multiple downstream tasks.
\end{itemize}
\section{Preliminary}

\myparagraph{Notations.} We define a text-attributed hypergraph (TAHG) as $\mathcal{H} = \{\mathcal{V}, \mathcal{E}, \mathcal{T}\}$, where $\mathcal{V} = \{v_i\}_{i=1}^{|\mathcal{V}|}$ denotes the set of nodes, and $\mathcal{E} = \{e_j\}_{j=1}^{|\mathcal{E}|}$ denotes the set of hyperedges, where each hyperedge $e \in \mathcal{E}$ is a non-empty subset of nodes, and \(|e|\) is the size of \(e\). Each node $v_i$ is associated with a textual sequence $t_i \in \mathcal{T}$, where $\mathcal{T}$ denotes the set of all text attributes. Each hyperedge $e_j$ is associated with a positive scalar weight $w_j$, and all weights formulate a diagonal matrix $\mathbf{W} \in \mathbb{R}^{|\mathcal{E}| \times |\mathcal{E}|}$.
The hypergraph structure can be modeled as an incidence matrix $\mathbf{H} \in \{0, 1\}^{|\mathcal{V}| \times |\mathcal{E}|}$, where $h_{ij} = 1$ if node $v_i$ belongs to hyperedge $e_j$, and $h_{ij} = 0$ otherwise. We define the node degree matrix $\mathbf{D}_v \in \mathbb{R}^{|\mathcal{V}| \times |\mathcal{V}|}$, where each entry is $d(v_i) = \sum_{e_j \in \mathcal{E}} w_j \cdot h_{ij}$. Similarly, the hyperedge degree matrix is defined as $\mathbf{D}_e \in \mathbb{R}^{|\mathcal{E}| \times |\mathcal{E}|}$, where each element $\delta(e_j) = \sum_{v_i \in e_j} h_{ij}$ represents the number of nodes connected by the hyperedge $e_j$.

\myparagraph{Hypergraph neural networks.}
HNNs have emerged as effective tools for representation learning on hypergraphs~\cite{HGNN,HCHA,AllSetTransformer}. Most HNNs adopt a two-stage message passing mechanism that alternates between node-to-hyperedge and hyperedge-to-node aggregation. At each layer $l$, the embedding of a hyperedge is updated based on its incident nodes, while a node embedding is updated via its associated hyperedges:
\begin{equation}
\begin{split}
    z^{(l)}_{e_j} & = f^{(l)}_{\mathcal{V} \rightarrow \mathcal{E}} \left( z^{(l-1)}_{e_j}, \{ z^{(l-1)}_{v_k} \mid v_k \in e_j \} \right), \\
    z^{(l)}_{v_i} & = f^{(l)}_{\mathcal{E} \rightarrow \mathcal{V}} \left( z^{(l-1)}_{v_i}, \{ z^{(l)}_{e_k} \mid v_i \in e_k \} \right),
\end{split}
\end{equation}
\noindent where $z^{(l)}_{e_j}$ and $z^{(l)}_{v_i}$ denote the embeddings of hyperedge $e_j$ and node $v_i$ at layer $l$, respectively. The functions $f^{(l)}_{\mathcal{V} \rightarrow \mathcal{E}}$ and $f^{(l)}_{\mathcal{E} \rightarrow \mathcal{V}}$ are permutation-invariant aggregation functions that propagate messages between nodes and hyperedges. This iterative process allows the model to capture complex higher-order relations in the hypergraph structure.

\myparagraph{Problem statement.} 
Given a TAHG $\mathcal{H} = \{\mathcal{V}, \mathcal{E}, \mathcal{T}\}$, the objective is to learn an optimal function $\Psi: \mathcal{H} \rightarrow (\mathbf{Z}_{\mathcal{V}}, \mathbf{Z}_{\mathcal{E}})$ in an unsupervised manner, mapping the hypergraph to low-dimensional node and hyperedge embeddings. These representations may then be applied to downstream tasks such as node classification and hyperedge prediction.
\section{Methodology}
\begin{figure*}[t]
\centering
\includegraphics[width=0.89\textwidth]{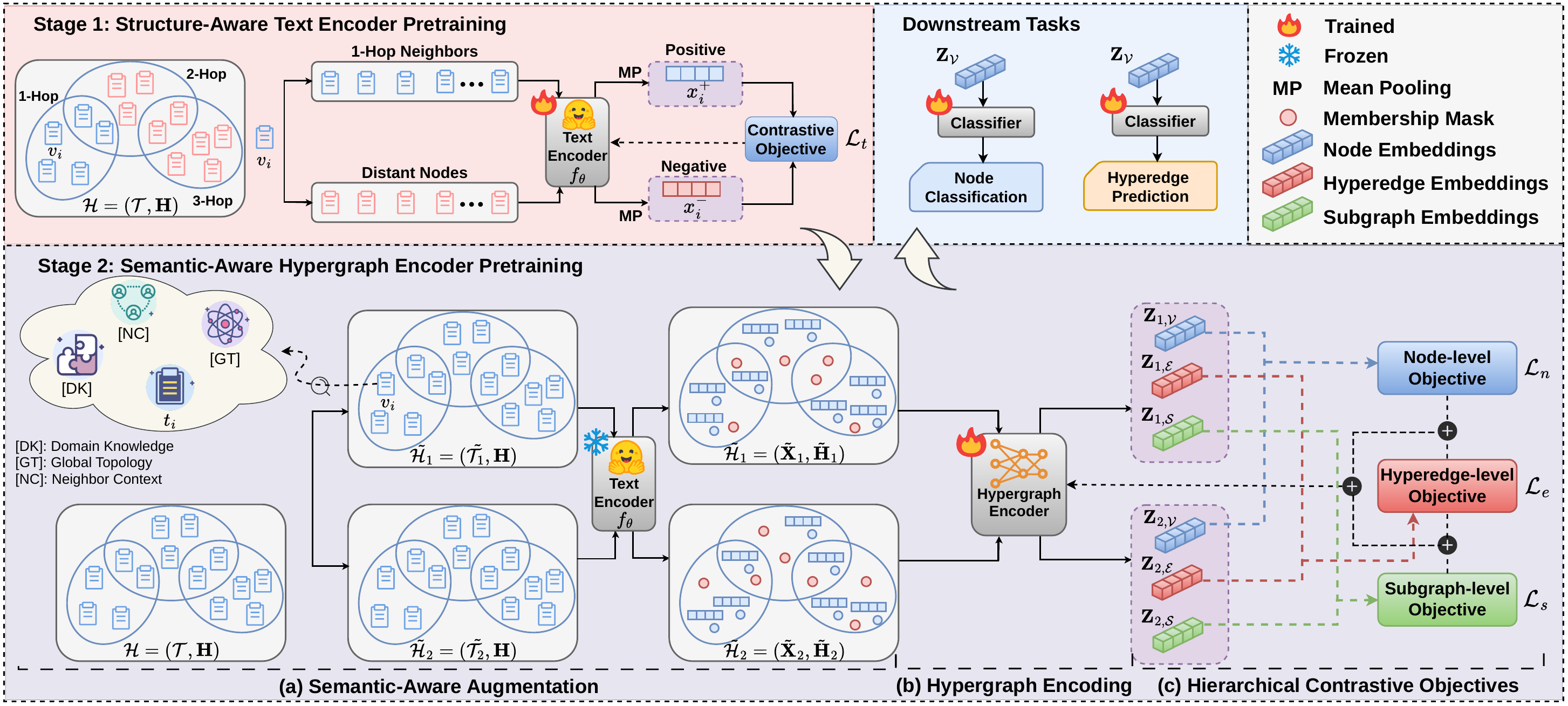} 
\caption{The overall framework of HiTeC. }
\vspace{-1em}
\label{overview}
\end{figure*}

In this section, we present HiTeC, a novel and effective SSL framework tailored for TAHGs. 
As illustrated in Figure~\ref{overview}, HiTeC adopts a two-stage paradigm.
In the first stage, a text encoder \(f_\theta\) is pre-trained with a structure-aware contrastive objective to capture topology-enhanced text representations.
In the second stage, with \(f_\theta\) frozen, we pre-train a hypergraph encoder \(g_\phi\) to fully exploit the rich semantic context of nodes through semantic-aware augmentations and hierarchical contrastive objectives at the node, hyperedge, and subgraph levels.

\subsection{Structure-aware Text Encoder Pretraining}
\myparagraph{Text encoder.}
Our text encoder \(f_{\theta}\) is built upon a pre-trained language model, followed by a two-layer MLP projection head that maps the [CLS] representation to a $d$-dimensional embedding.
Given a node \(v_i\) with textual attribute \(t_i\), the resulting text embedding \(x_i \in \mathbb{R}^d\) is computed as:
\begin{equation}
x_i = f_{\theta}(t_i).
\end{equation}

\myparagraph{Structure-aware contrastive objective.}
In real-world hypergraphs, nodes are often semantically closer to their local neighbors than to distant ones. To encode this structural prior, we treat the 1-hop neighbors \(N^{(1)}_{v_i}\) of \(v_i\), i.e., nodes sharing at least one hyperedge with \(v_i\), as positives, while the remaining nodes \(\mathcal{V} \setminus N^{(1)}_{v_i}\) are treated as negatives. Positive and negative representations are constructed by mean pooling the corresponding text embeddings:
\begin{equation}
\begin{split}
x_i^+ &= \frac{1}{|N^{(1)}_{v_i}|} \sum_{v_j \in N^{(1)}_{v_i}} x_j, \\
x_i^- &= \frac{1}{|\mathcal{V} \setminus N^{(1)}_{v_i}|} \sum_{v_j \in \mathcal{V} \setminus N^{(1)}_{v_i}} x_j.
\end{split}
\end{equation}

A triplet margin loss~\cite{triplet2015} is adopted to guide \(f_\theta\) in distinguishing positive sample \( x_i^+ \) from negative sample \( x_i^- \).
For each anchor \(x_i\), the loss is defined as:
\begin{equation}
\ell_t(x_i, x_i^+, x_i^-)
=
\max
\big(
\cos(x_i, x_i^-)
-
\cos(x_i, x_i^+)
+
m,
0
\big),
\end{equation}
where \( \cos(\cdot) \) denotes cosine similarity, and \( m \) is the margin hyperparameter. The final objective is:
\begin{equation}
\mathcal{L}_t =
\frac{1}{|\mathcal{V}|}
\sum_{v_i \in \mathcal{V}}
\ell_t(x_i, x_i^+, x_i^-).
\end{equation}

\subsection{Semantic-aware Hypergraph Encoder Pretraining}
Existing HCL methods typically overlook node textual semantics and rely on random augmentation strategies, resulting in suboptimal performance~\cite{gaugllm2024,ctaug2024}. 
Moreover, they struggle to model long-range dependencies, limiting broader contextual understanding~\cite {hash-code2024,ctaug2024}.
To tackle these issues, we propose a semantic-aware contrastive learning strategy to pre-train \( g_{\phi} \). As shown in Figure~\ref{overview}, this stage comprises three components: semantic-aware augmentation, hypergraph encoding, and hierarchical contrastive objectives. 

\myparagraph{Semantic-aware augmentation.}
To eliminate the impact of noise arising from random perturbation, we introduce two semantic-aware augmentation strategies: structure-contextualized text augmentation and semantic-aware hyperedge dropping.

\noindent\underline{\textit{Structure-contextualized text augmentation.}}
As perturbations applied directly to text embeddings can easily distort semantic meaning, we instead augment raw text with structure-aware contextual information. Specifically, for each node text attribute \(t_i\), we first construct a structure-contextualized text view by concatenating three contextual components with the original text:
\(\tilde{t}_{1,i} =
[\texttt{[DK]} \,||\, \texttt{[GT]} \,||\, \texttt{[NC]} \,||\, t_i]\),
where:
(i) \texttt{[DK]} denotes domain knowledge (e.g., academic or e-commerce) shared across all nodes;
(ii) \texttt{[GT]} encodes global topological information for node \(v_i\) (e.g., node degree and local sparsity);
and (iii) \texttt{[NC]} represents neighbor context, constructed by concatenating the raw texts of sampled neighbors of \(v_i\).
The second text view is the original text, i.e., \(\tilde{t}_{2,i} = t_i\).
This simple yet effective strategy generates semantically consistent but contextually diverse text variants, which are encoded by the shared encoder \(f_\theta\) into a unified embedding space, denoted as \( \tilde{\mathbf{X}}_1 \) and \( \tilde{\mathbf{X}}_2 \), respectively.
The full template for constructing the structure-contextualized text view is provided in Appendix~\ref{sec:template}.

\noindent\underline{\textit{Semantic-aware hyperedge dropping.}}
Effective structure augmentation should consider both textual semantics and topology. To achieve this, we define a semantic cohesiveness score that reflects both the structural grouping of nodes and their semantic similarity, which is further used to guide the drop probability of each node--hyperedge incidence. Connections within more cohesive hyperedges are more likely to be preserved, encouraging the retention of structurally and semantically aligned patterns, while selectively pruning noisy or weakly node–hyperedge connections.
\begin{definition}[\emph{\textbf{semantic cohesiveness score}}]
Given a TAHG \( \mathcal{H}(\mathcal{V}, \mathcal{E}, \mathbf{X}) \), the semantic cohesiveness score of a hyperedge \( e \in \mathcal{E} \) is defined as the average pairwise similarity among its member nodes. Let \( \mathbf{X}_e \in \mathbb{R}^{|e| \times d} \) denote the normalized feature matrix for nodes in \( e \). The semantic cohesiveness score \( s(e) \) is then computed as:
\begin{equation*}
s(e) = \frac{1}{\binom{|e|}{2}} \sum_{i=1}^{|e|} \sum_{j=i+1}^{|e|} \bm{S}_e(i, j), \quad
\bm{S}_e = \mathbf{X}_e \mathbf{X}_e^\top,
\end{equation*}
where \( \bm{S}_e(i, j) \) denotes the similarity between the \( i \)-th and \( j \)-th nodes in hyperedge \( e \).
\end{definition}

The drop probability of each incidence \(  (v_i, e_j)\), where \(v_i \in e_j \), is then determined by \(s(e_j)\):
\begin{equation}
p_{\text{drop}}(v_i, e_j) = 1 - \sigma \left( \frac{s(e_j) - 0.5}{\tau_{\text{drop}}} \right),
\end{equation}
where \( \sigma(\cdot)\) is the sigmoid function and \( \tau_{\text{drop}} \) is a temperature hyperparameter. 
Higher cohesiveness scores correspond to lower drop probabilities. In implementation, we construct a binary masking matrix \( \mathbf{M} \in \{0, 1\}^{|\mathcal{V}| \times |\mathcal{E}|} \), where entries \( m_{ij} \sim \mathcal{B}(1 - p_{\text{drop}}(v_i, e_j)) \) are sampled from a Bernoulli distribution. The perturbed matrix \( \tilde{\mathbf{H}} \) is:
\begin{equation}
\tilde{\mathbf{H}} = \mathbf{M} \odot \mathbf{H},
\end{equation}
where \( \odot \) denotes element-wise multiplication. We apply this semantic-aware hyperedge drop mechanism to both views, yielding two augmented hypergraphs: \( \tilde{\mathcal{H}}_1 = (\tilde{\mathbf{X}}_1, \tilde{\mathbf{H}}_1) \) and \( \tilde{\mathcal{H}}_2 = (\tilde{\mathbf{X}}_2, \tilde{\mathbf{H}}_2) \). This procedure complements text-level augmentation by introducing structural diversity while retaining semantic consistency, enabling the hypergraph encoder to learn more robust node representations.

\myparagraph{Hypergraph encoder.}
We take the HGNN~\cite{hgnn2019} with the element-wise mean pooling layer as our hypergraph encoder. 
At the \( l \)-th layer, the hyperedge representation \( \mathbf{Z}_{\mathcal{E}}^{(l)} \in \mathbb{R}^{|\mathcal{E}| \times d} \) and the node representation \( \mathbf{Z}_{\mathcal{V}}^{(l)} \in \mathbb{R}^{|\mathcal{V}| \times d} \) are updated as:
\begin{equation}
\begin{split}
\mathbf{Z}_{\mathcal{E}}^{(l)} &=  \rho\left( \mathbf{D}_e^{-1} \mathbf{H}^\top \mathbf{Z}_{\mathcal{V}}^{(l-1)} \mathbf{\Theta}_{\mathcal{E}}^{(l)} + \bm{b}_e^{(l)}  \right), \\
\mathbf{Z}_{\mathcal{V}}^{(l)} &= \rho\left( \mathbf{D}_v^{-1} \mathbf{H} \mathbf{W} \mathbf{Z}_{\mathcal{E}}^{(l)} \mathbf{\Theta}_{\mathcal{V}}^{(l)} + \bm{b}_v^{(l)} \right),
\end{split}
\end{equation}
where \( \rho(\cdot) \) is the PReLU activation function, \( \mathbf{\Theta}_{\mathcal{E}}^{(l)} \) and \( \mathbf{\Theta}_{\mathcal{V}}^{(l)} \) are trainable weight matrices, and \( \bm{b}_e^{(l)} \) and \( \bm{b}_v^{(l)} \) are 
bias terms. \( \mathbf{W} \) is initialized as an identity matrix, assigning equal weights to all hyperedges.

\myparagraph{Hierarchical contrastive objectives.}
Prior works~\cite{tricl2023,se-hssl2024} have shown that multi-scale contrastive objectives are beneficial for HSSL. To better capture structural semantics at different granularities, we adopt a hierarchical contrastive learning framework across node, hyperedge, and subgraph levels. In particular, we introduce a novel subgraph-level objective based on $s$-walk sampling to capture long-range dependencies beyond local structures. All objectives are optimized using the widely adopted contrastive loss, InfoNCE~\cite{infonce2018}.

\noindent\underline{\textit{Node-level objective.}}
We first model the correlation of nodes across two augmented views. Given a node \( v_i \in \mathcal{V} \), let \( \mathbf{z}_{1, v_i} \) and \( \mathbf{z}_{2, v_i} \) denote its node embeddings derived from two augmented hypergraphs. 
 By fixing \(\mathbf{z}_{1,v_i} \) as anchor, we treat\( (\mathbf{z}_{1,v_i}, \mathbf{z}_{2,v_i}) \) as a positive pair, and all \( \mathbf{z}_{2,v_k} \) for \( k \neq i \) as negative samples.
The loss of anchor \(\mathbf{z}_{1,v_i} \) is:
\begin{equation}
    \ell_n(\mathbf{z}_{1,v_i}, \mathbf{z}_{2,v_i}) = -\log \frac{e^{\cos(\mathbf{z}_{1,v_i}, \mathbf{z}_{2,v_i})/\tau_n}}{\sum_{k=1}^{|\mathcal{V}|} e^{\cos(\mathbf{z}_{1,v_i}, \mathbf{z}_{2,v_k})/\tau_n}},
\end{equation}
where \( \tau_n \) is the temperature coefficient. In practice, we further symmetrize the contrastive loss by using the second view's embedding as the anchor, yielding the final node-level contrastive objective:
\begin{equation}
\label{node_loss}
\mathcal{L}_n = \frac{1}{2|\mathcal{V}|} \sum_{i=1}^{|\mathcal{V}|} \left[ \ell_n(\mathbf{z}_{1,v_i}, \mathbf{z}_{2,v_i}) + \ell_n(\mathbf{z}_{2,v_i}, \mathbf{z}_{1,v_i}) \right].
\end{equation}

\noindent\underline{\textit{Hyperedge-level objective.}}
To consider group-level knowledge, we integrate hyperedge-level contrastive signal during optimization.
Given a hyperedge \( e_j \in \mathcal{E} \), let \( \mathbf{z}_{1,e_j} \) and \( \mathbf{z}_{2, e_j} \) be its hyperedge representations from the two augmented views. Similar to the node-level objective, the contrastive loss with \(\mathbf{z}_{1,e_j} \) as anchor is formulated as:
\begin{equation}
    \ell_{\text{e}}(\mathbf{z}_{1,e_j}, \mathbf{z}_{2,e_j}) = -\log \frac{e^{\cos(\mathbf{z}_{1,e_j}, \mathbf{z}_{2,e_j})/\tau_e}}{\sum_{k=1}^{|\mathcal{E}|} e^{\cos(\mathbf{z}_{1,e_j}, \mathbf{z}_{2,e_k})/\tau_e}},
\end{equation}
where \( \tau_e \) is the temperature coefficient. 
By symmetrizing the contrastive loss, the final hyperedge-level objective can be written as:
\begin{equation}
\label{edge_loss}
    \mathcal{L}_{e} = \frac{1}{2|\mathcal{E}|} \sum_{j=1}^{|\mathcal{E}|} \left[ \ell_e(\mathbf{z}_{1,e_j}, \mathbf{z}_{2,e_j}) + \ell_e(\mathbf{z}_{2,e_j},\mathbf{z}_{1,e_j}) \right].
\end{equation}

\noindent\underline{\textit{Subgraph-level objective.}}
\begin{figure}[t]
\centering
\includegraphics[width=1\columnwidth]{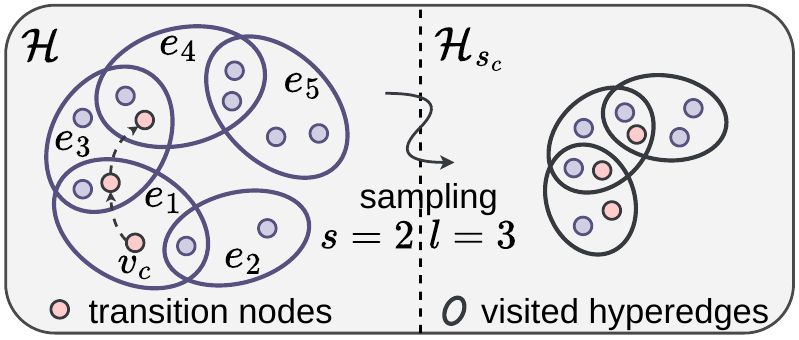} %
\caption{An example of the \(s\)-walk-based subgraph sampling process. Given the center node \(v_c\), we perform an \(s\)-walk with \( s = 2 \) and \( l = 3 \) on \(\mathcal{H}\). The resulting hyperedge traversal sequence is \([e_1, e_3, e_4]\), and the sampled subgraph is denoted as \( \mathcal{H}_{s_c} \).}
\vspace{-0.5em}
\label{s-walk}
\end{figure}
While node- and hyperedge-level objectives effectively capture local structural signals, they remain insufficient for modeling long-range dependencies in hypergraphs. We therefore introduce a subgraph-level objective based on an \(s\)-walk sampling strategy. Unlike random walks, an \(s\)-walk defines a high-order traversal sequence in hypergraphs, where \(s\) controls the overlap strength between consecutive hyperedges, enabling the extraction of cohesive and semantically meaningful substructures unique to hypergraphs~\cite{s-walk2020,s-walk2024}. We next formally define \(s\)-walk and describe our subgraph sampling method in detail.

\begin{definition}[\emph{\textbf{\(s\)-walk}}]
Given \( s \in \mathbb{N}^+ \), an \(s\)-walk of length \( l \) is defined as an ordered sequence of hyperedges \( [e_1, \ldots, e_l] \), where every consecutive pair of hyperedges shares at least \( s \) common nodes, i.e., \( |e_j \cap e_{j+1}| \geq s \) for all \( j \in [1, l-1] \).
\end{definition}

As illustrated in Figure~\ref{s-walk}, an \(s\)-walk-based subgraph
\( \mathcal{H}_{s_c} = (\mathcal{V}_{s_c}, \mathcal{E}_{s_c}) \)
is constructed from a center node \(v_c\) via iterative \(s\)-walk sampling. 
We first randomly select a hyperedge \(e_1\) containing \(v_c\). At each step, we identify hyperedges sharing at least \(s\) nodes with the current hyperedge \(e_j\), randomly sample one as \(e_{j+1}\), and then randomly choose a node from \(e_{j+1}\) as the transition node. The process stops when the traversal length reaches \(l\). 
The sampled subgraph consists of the visited hyperedges
\( \mathcal{E}_{s_c} = \{e_j\}_{j=1}^{l} \)
and their associated nodes
\( \mathcal{V}_{s_c} = \bigcup_{j=1}^{l} e_j \).
Each sampled subgraph \( \mathcal{H}_{s_c} \) is encoded by the shared hypergraph encoder \(g_{\phi}(\cdot)\), and its representation is obtained through mean pooling \(P_{\text{mean}}\) over the corresponding node embeddings:
\begin{equation}
\mathbf{z}_{s_c}
= P_{\text{mean}}\!\left( g_{\phi}(\mathcal{H}_{s_c}) \right).
\end{equation}

However, constructing and encoding subgraphs for all nodes incurs a high computational cost. 
To mitigate this, we follow prior findings suggesting that high-degree nodes tend to carry richer contextual information~\cite{hyperkan2025}.
Specifically, we select a representative node subset \( \mathcal{V}_r \subseteq \mathcal{V} \) by ranking node degrees and retaining the top \( r\% \). As validated in our experiments, this strategy maintains competitive performance even with a significantly reduced number of subgraphs. 
For each anchor node \( v_m \in \mathcal{V}_r \), we obtain subgraph representations \( \mathbf{z}_{1,s_m} \) and \( \mathbf{z}_{2,s_m} \) from two augmented views. Similar to the former contrastive signals, the subgraph-level loss with anchor \( \mathbf{z}_{1,s_m} \) is defined as:
\begin{equation}
    \ell_{\text{s}}(\mathbf{z}_{1,s_m},\mathbf{z}_{2,s_m}) = -\log \frac{ e^{\cos(\mathbf{z}_{1,s_m}, \mathbf{z}_{2,s_m})/\tau_s}}{\sum_{k=1}^{|\mathcal{V}_r|} e^{\cos(\mathbf{z}_{1,s_m}, \mathbf{z}_{2,s_k})/\tau_s}},
\end{equation}
where \( \tau_s \) is a temperature coefficient. And the final symmetric loss across all anchors is:
\begin{equation}
\label{subgraph_loss}
  \mathcal{L}_{\text{s}} = \frac{1}{2|\mathcal{V}_r|} \sum_{m=1}^{|\mathcal{V}_r|} \left[ \ell_s(\mathbf{z}_{1,s_m}, \mathbf{z}_{2,s_m}) + \ell_{\text{s}}(\mathbf{z}_{2,s_m}, \mathbf{z}_{1,s_m}) \right].
\end{equation}

\noindent\underline{\textit{Overall objective.}}
Combining Eq.~(\ref{node_loss}), (\ref{edge_loss}), and (\ref{subgraph_loss}), the overall contrastive loss is defined as: contrastive loss is defined as:
\begin{equation}
\mathcal{L}
= \mathcal{L}_{n}
+ \lambda_e \mathcal{L}_e
+ \lambda_s \mathcal{L}_s ,
\end{equation}
where \( \lambda_e \) and \( \lambda_s \) are balancing weights. A detailed complexity analysis is provided in Appendix~\ref{app:complexity}.



\section{Experiments}
\label{exp}
\begin{table}[t]
  \centering
  \small
  \begin{adjustbox}{width=\linewidth}
    \begin{tabular}{cccccc}
    \toprule
    \textbf{Dataset} & \textbf{\(|\mathcal{V}|\)} & \textbf{\( |\mathcal{E}| \)}  & \textbf{ Avg.\(|e|\)} & \textbf{\# Avg.tokens} & \textbf{\# Class}\\
    \midrule
    Citeseer  & 1,778   & 2,118    & 2 & 198 & 6 \\
    Cora      & 2,708   & 1,579     & 3  & 189 & 7 \\
    History   & 41,551  & 169,454   & 9  & 300 & 12 \\
    Photo     & 48,362  & 212,247   & 11 & 189 & 12 \\
    Computers & 87,229  & 277,539   & 9  & 115 & 10 \\
    Fitness   & 173,055 & 1,468,229  & 11 & 28  & 13 \\
    \bottomrule
    \end{tabular}
  \end{adjustbox}
  \caption{\label{dataset}
    Statistics of datasets. 
  }
  \vspace{-1em}
\end{table}
\begin{table*}[ht]
\centering
\footnotesize 
\renewcommand{\arraystretch}{0.9} 
\setlength{\tabcolsep}{4pt}
\begin{tabular}{llcccccc}
\toprule
\textbf{} & \textbf{Method} & \textbf{Citeseer} & \textbf{Cora} & \textbf{History} & \textbf{Photo} & \textbf{Computers} & \textbf{Fitness} \\
\midrule
\midrule
\multirow{6}{*}{\rotatebox{90}{\textbf{Graph SSL}}}
&GraphCL + SE          & 59.63 ± 1.58 & 56.69 ± 1.99 & 62.86 ± 0.22 & 42.61 ± 0.21 & 34.45 ± 0.38 & 52.11 ± 0.43 \\
&GraphCL + GIANT       & 57.58 ± 1.90 & 55.64 ± 1.50 & 62.08 ± 0.24 & 42.74 ± 0.22 & 34.65 ± 0.23 & 52.04 ± 0.34 \\
&BGRL + SE             & 60.81 ± 1.45 & 59.72 ± 1.32 &  65.74 ± 0.30 & 44.60 ± 0.30 & 39.40 ± 0.46 & 55.66 ± 0.33 \\
&BGRL + GIANT          & 58.68 ± 1.92 & 57.44 ± 1.39 & 63.99 ± 0.35 & 43.65 ± 0.25 & 37.00 ± 0.37 & 55.05 ± 0.25 \\
&GraphMAE + SE         & 46.42 ± 2.75 & 50.33 ± 0.47 & 52.90 ± 9.01 & 41.52 ± 0.08 & 25.61 ± 0.11 & 40.65 ± 0.37 \\
&GraphMAE + GIANT      & 50.11 ± 1.74 & 53.59 ± 1.07 & 56.20 ± 0.12 & 41.54 ± 0.06 & 25.57 ± 0.12 & 42.04 ± 0.04 \\
\midrule
\multirow{11}{*}{\rotatebox{90}{\textbf{HSSL}}}
& TriCL + SE         & 60.78 ± 1.00 & 58.79 ± 1.30  & 69.28 ± 0.32 & 47.11 ± 0.44  & 43.43 ± 0.66  & 45.46 ± 0.21\\
& TriCL + BERT       & 64.50 ± 1.08 & 62.67 ± 1.13  & \underline{75.01 ± 0.22} & 51.25 ± 0.62  & \underline{46.60 ± 0.90} & 48.97 ± 0.63\\
& TriCL + RoBERTa    & 64.48 ± 0.95 & 60.74 ± 1.01  & 70.62 ± 0.81 & 48.42 ± 0.67  & 43.82 ± 1.11 & 47.49 ± 0.60 \\
& SE-HSSL + SE       & 57.04 ± 3.31 & 59.87 ± 1.83 & 68.59 ± 0.74 & 43.75 ± 2.01 & 36.53 ± 2.32 & 54.67 ± 2.00\\
&SE-HSSL + BERT      & \underline{65.83 ± 1.21} & \underline{64.30 ± 0.89}  & 74.88 ± 0.42 & \underline{51.80 ± 0.63} & 41.63 ± 2.21 & \underline{56.92 ± 1.44}\\
&SE-HSSL + RoBERTa   & 65.04 ± 1.04 & 62.76 ± 1.15 & 72.13 ± 1.07 & 49.41 ± 2.62 & 40.86 ± 3.99 & 54.79 ± 0.64\\
&HypeBoy + SE        & 59.98 ± 2.45 & 29.87 ± 0.62 & OOM & OOM & OOM & OOM \\
&HypeBoy + BERT      & 61.98 ± 1.70 & 32.73 ± 1.71 & OOM & OOM & OOM & OOM \\
&HypeBoy + RoBERTa   & 50.41 ± 2.88 & 30.84 ± 0.95 & OOM & OOM & OOM & OOM \\
&VilLain (w/o feat.)  & 42.22 ± 1.72 & 53.72 ± 1.00 & 40.99 ± 3.06 & 33.94 ± 1.39  & 20.99 ± 0.91 & 42.04 ± 0.04\\
&HyperBERT           & 58.25 ± 3.04 & 47.03 ± 0.77 & 62.12 ± 0.19 & 41.46 ± 0.14 & 26.79 ± 0.20 & OOM \\
\cmidrule(lr){2-8}
&HiTeC               & \textbf{68.69 ± 0.97} & \textbf{74.07 ± 0.88} & \textbf{79.81 ± 0.16} & \textbf{59.64 ± 0.16} & \textbf{55.81 ± 0.12} & \textbf{67.18 ± 0.19} \\
\bottomrule
\end{tabular}
\caption{Evaluation results on node classification task(\% mean \(\pm\) std). The \textbf{best} and \underline{second-best} results are highlighted in \textbf{bold} and \underline{underline}, respectively. ``w/o feat.'' denotes methods without node features. OOM indicates that the method runs out of memory on a single 80GB GPU.}
\label{tab:node}
\vspace{-0.5em}
\end{table*}

\subsection{Experimental Setup}
\myparagraph{Datasets.}
As no public TAHG datasets are available, we construct six text-attributed hypergraph datasets via max clique-based hypergraph reconstruction~\cite{maxclique1973,reconstruction2024}.
They include two co-citation hypergraphs (Citeseer~\cite{citeseer} and Cora~\cite{cora}) and four co-purchasing networks (History, Photo, Computers, and Fitness~\cite{cs-tag2023}).
Dataset statistics are summarized in Table~\ref{dataset}, with details in Appendix~\ref{app:dataset}.

\myparagraph{Baselines.}
We compare HiTeC with eight representative SSL baselines:
(1) three graph-based methods (GraphCL~\cite{graphcl2020}, BGRL~\cite{bgrl2022}, and GraphMAE~\cite{graphmae2022});
(2) four hypergraph-based methods (TriCL~\cite{tricl2023}, SE-HSSL~\cite{se-hssl2024}, HypeBoy~\cite{hypeboy2024}, and VilLain~\cite{villain2024});
and (3) one TAHG-based method, HyperBERT~\cite{hyperbert2024}.
Since graph- and hypergraph-based methods cannot process raw text directly, we extract node features using different text encoders. Specifically, both graph- and hypergraph-based methods use shallow encoders (SE), i.e., Skip-Gram~\cite{skip2013}. Graph-based methods additionally use GIANT~\cite{giant2022}, a self-supervised encoder for text-attributed graphs (TAGs), while hypergraph-based methods use pretrained language models, including BERT~\cite{bert2019} and RoBERTa~\cite{roberta2019}. Detailed descriptions of the baselines are provided in Appendix~\ref{app:baseline}.

\myparagraph{Evaluation protocol.}
We evaluate our approach on two hypergraph benchmark tasks: node classification and hyperedge prediction. After pretraining with HiTeC, we adopt the linear evaluation protocol~\cite{tricl2023,se-hssl2024,hypeboy2024}, where the learned node embeddings are frozen and used as input features for downstream classifiers. 
For node classification, we train a logistic regression classifier with a 10\%/10\%/80\% train/validation/test split. 
For hyperedge prediction, following~\cite{hypeboy2024}, we sample an equal number of negative hyperedges to match the ground-truth positives and split the data using a 60\%/20\%/20\% ratio, after which a two-layer MLP classifier is trained for prediction.
For each dataset, we report the average test accuracy and standard deviation over 20 random data splits, each evaluated with 5 different random initializations. 
Full evaluation details are provided in Appendix~\ref{app:ep}.

\myparagraph{Implementation details.}
For HiTeC, we adopt BERT-Base~\cite{bert2019} as the default text encoder and apply LoRA~\cite{lora2022} for parameter-efficient fine-tuning. We further evaluate other text encoder backbones in Appendix~\ref{app:backbones}. We tune the \(s\)-walk length \( l \) and overlap size \( s \) in \(\{1,2,3,4,5,10\}\), the sampling ratio \( r\) in \(\{10,20,30,40,50\}\), and \( \lambda_e \) and \( \lambda_s \) in \(\{1,2,3,4\}\). Additional implementation details are shown in Appendix~\ref{app:imp}. 

\subsection{Main Results}
In this section, we evaluate our approach on two representative hypergraph tasks: node classification and hyperedge prediction (see Appendix~\ref{app:hp}).

\myparagraph{Node classification evaluation.}
Table~\ref{tab:node} presents the node classification results, revealing two key observations:
(1) Graph-based SSL methods consistently underperform HSSL methods due to the loss of high-order relations caused by clique expansion.
Even the TAG encoder GIANT, which jointly encodes text and pairwise structure, remains insufficient for modeling higher-order structures, highlighting the need for dedicated SSL methods for TAHGs.
(2) HiTeC achieves the best overall performance across all datasets.
Methods that neglect rich node features (e.g., VilLain) achieve limited performance, highlighting the importance of textual semantics in TAHG learning.
HCL methods (e.g., TriCL and SE-HSSL) rely on structure-driven contrastive signals without semantic-aware augmentation, resulting in suboptimal performance.
While HyperBERT jointly models textual and hypergraph information, it lacks multi-scale structural and semantic contrastive signals.
In contrast, HiTeC consistently outperforms all baselines, surpassing the second-best method by 9.77\% and 10.26\% on Cora and Fitness, respectively.
These improvements are mainly attributed to semantic-aware augmentations, which reduce perturbation noise, and the \(s\)-walk-based subgraph signals, which capture more informative high-order dependencies.
Additional analysis on efficiency is provided in Appendix~\ref{app:efficiency}.

\subsection{Ablation Study}
\begin{figure}[t!]
  \centering
  \begin{subfigure}[b]{0.48\linewidth}
    \centering
    \includegraphics[width=\linewidth]{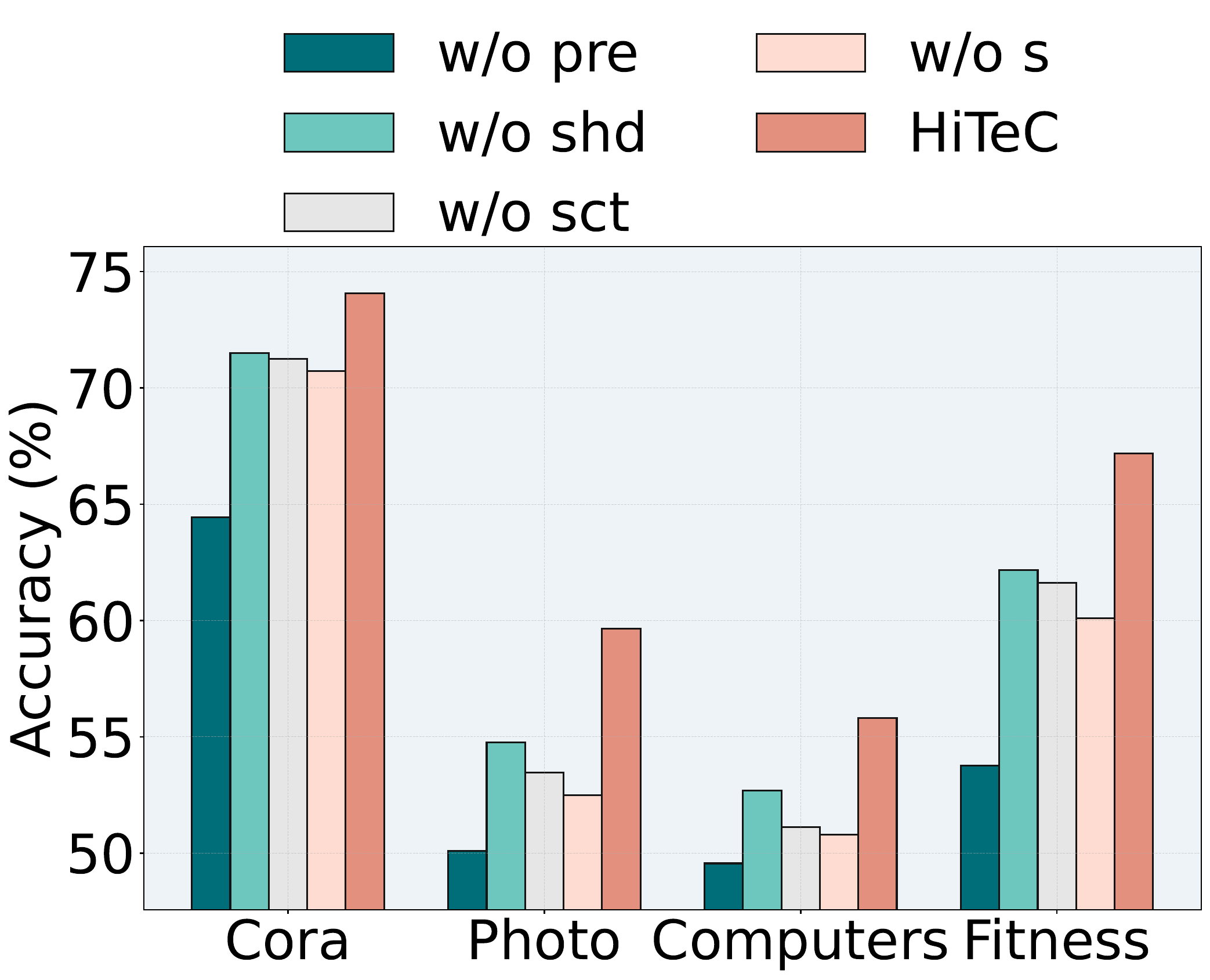}
    \caption{Key component ablation.}
    \label{ab:components}
  \end{subfigure}
  \hfill
  \begin{subfigure}[b]{0.48\linewidth}
    \centering
    \includegraphics[width=\linewidth]{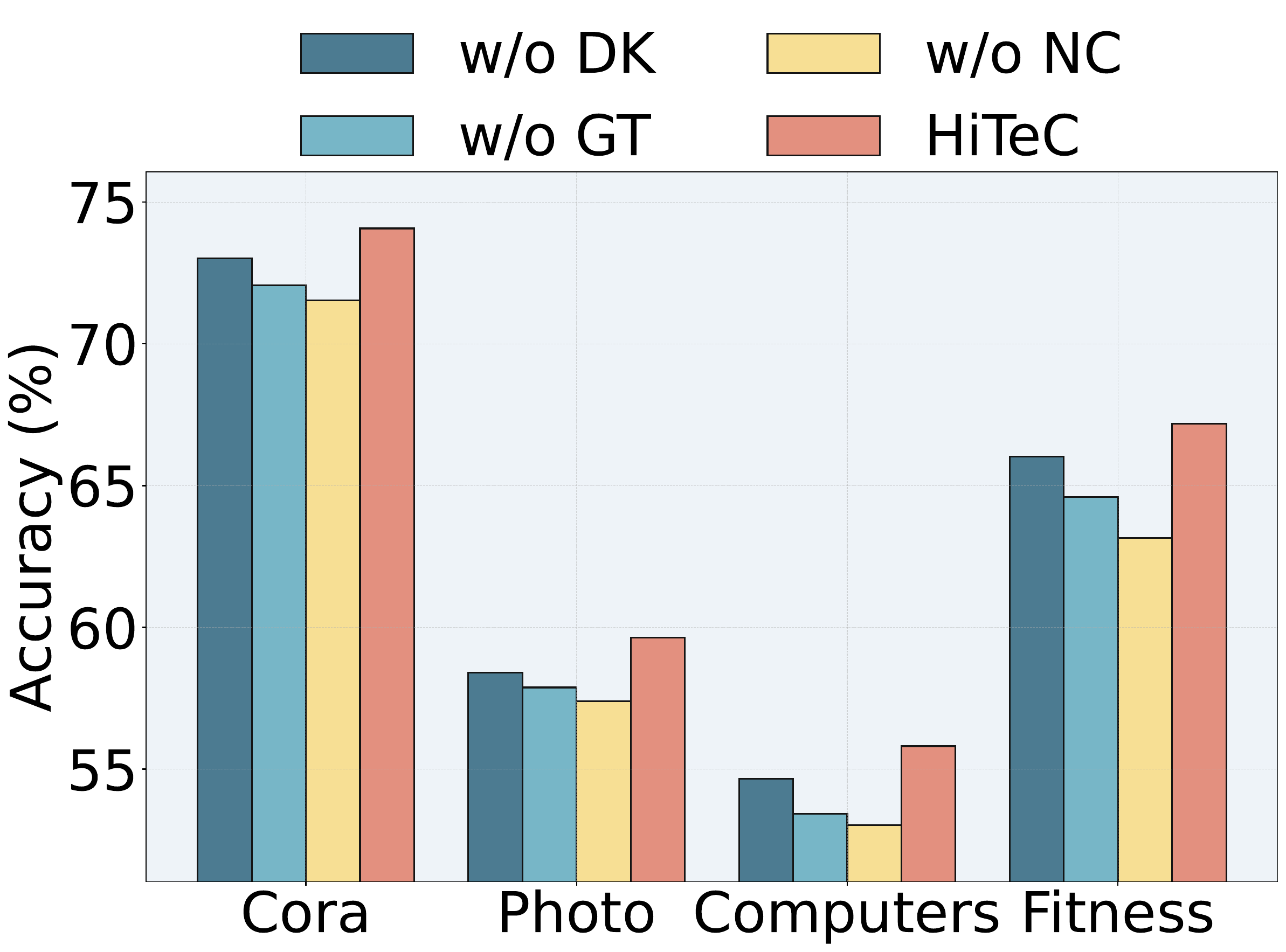}
    \caption{Text view ablation.}
    \label{ab:text-view}
  \end{subfigure}
  \caption{Ablation study. (a) Key component ablation: ``w/o pre'' removes structure-aware text encoder pre-training; ``w/o shd'' replaces semantic-aware hyperedge dropping with random drop; ``w/o sct'' removes the structure-contextualized text view; and ``w/o s'' removes the subgraph-level objective. (b) Text view ablation: ``w/o DK'', ``w/o GT'', and ``w/o NC'' remove domain knowledge, global topology, and neighbor context from the structure-contextualized text view, respectively.}
  \vspace{-0.5em}
  \label{fig:ablation}
\end{figure}

In this section, we conduct ablation studies on: (1) key components of HiTeC, (2)  structure-contextualized text view components.


\myparagraph{Impact of key components.}
We evaluate the contribution of key components in HiTeC. 
As shown in Figure~\ref{fig:ablation}(a), removing any module noticeably degrades performance. 
The most significant drop is observed when removing text encoder pre-training (`w/o pre''), resulting in a decrease of 13.41\% on Fitness, highlighting the effectiveness of injecting topological information into the text encoder. 
The subgraph-level objective also plays a critical role: removing it (`w/o s'') causes a decrease of 7.14\% on Photo, demonstrating the benefit of modeling long-range dependencies inherent in TAHGs.

\myparagraph{Impact of structure-contextualized text view components.}
We further evaluate the effectiveness of different components in the structure-contextualized text view in Figure~\ref{fig:ablation}(b). 
Overall, the complete view consistently outperforms all alternative variants. Specifically, removing neighbor context (``w/o NC'') and global topology (``w/o GT'') leads to pronounced performance degradation, underscoring the importance of incorporating meaningful structural signals into the original text.


\subsection{Parameter Sensitivity Analysis}
We investigate the impact of three key hyperparameters: the overlap size $s$, the subgraph sampling ratio $r$, and the walk length $l$ (see Appendix~\ref{app:length}).
\begin{figure}[t]
\centering
\includegraphics[width=1\columnwidth]{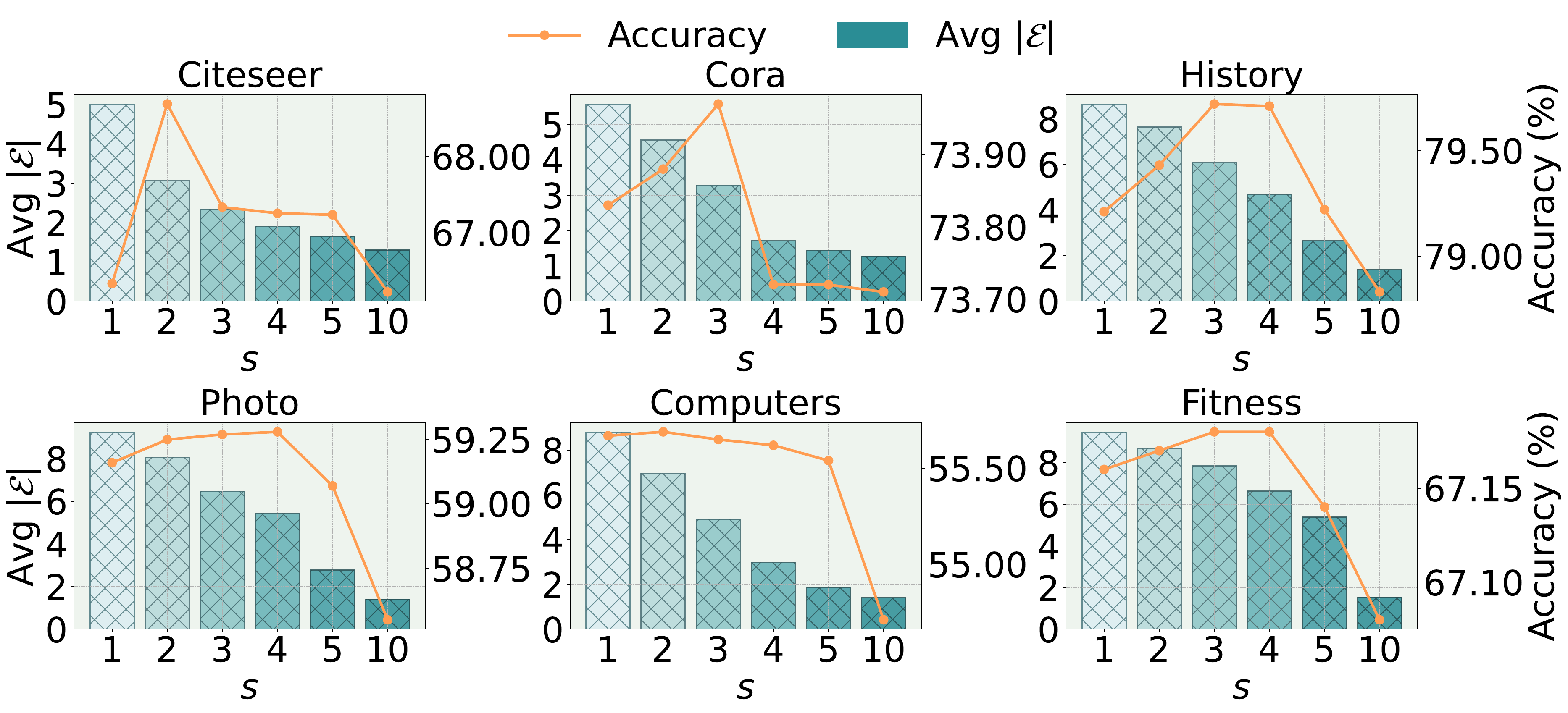} 
\caption{
Sensitivity analysis of overlap size $s$. Avg$|\mathcal{E}|$ indicates the average number of sampled hyperedges.
}
\vspace{-1em}
\label{s_param}
\end{figure}

\myparagraph{Overlap size $s$.}
Figure~\ref{s_param} illustrates how model performance varies with different \(s\)  while fixing \(l=4\). 
Model accuracy first improves and then degrades as \(s\) increases, reflecting a trade-off between structural coherence and exploration range. Moderate overlap constraints (e.g., \(s=3\)) encourage cohesive subgraphs while preserving contextual diversity, leading to the best performance in most cases. In contrast, \(s=1\) reduces to a vanilla random walk with weak structural correlation, while overly large overlap constraints (e.g., \(s=10\)) greatly restrict candidate hyperedges, reducing contextual diversity and impairing performance.

\myparagraph{Subgraph sampling ratio $r$.}
\begin{figure}[t]
\centering
\includegraphics[width=1\columnwidth]{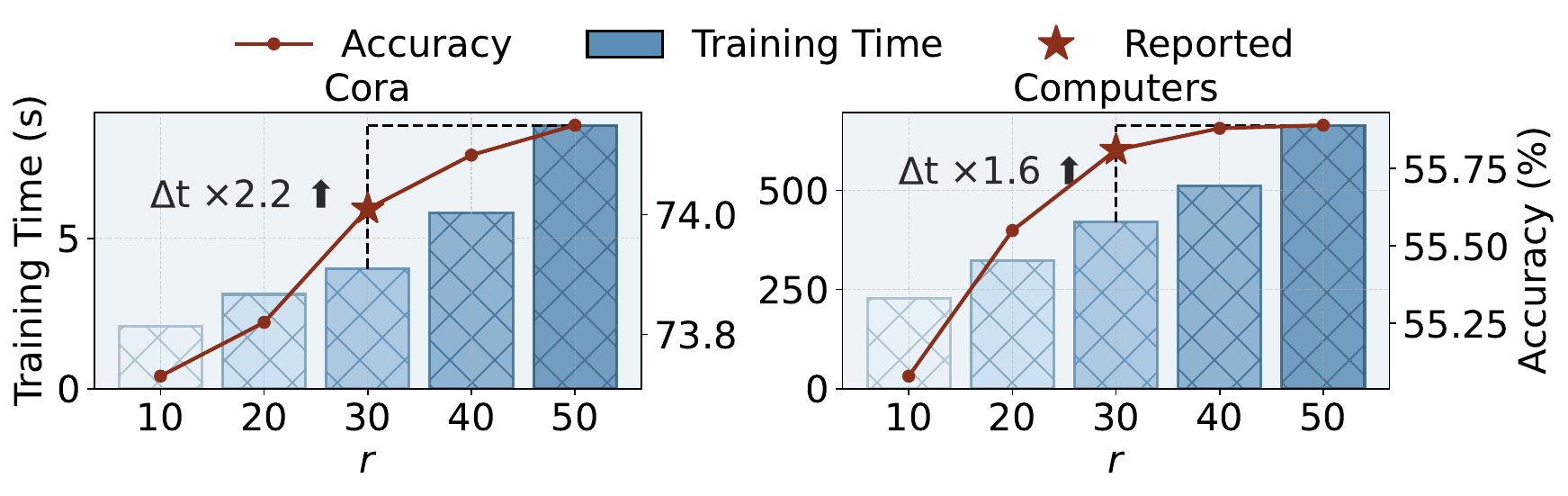} 
\caption{
Training time and accuracy under different sampling ratios \(r\).
\(\Delta t\) indicates the relative speedup.
}
 \vspace{-1em}
\label{fig:sampling}
\end{figure}
We analyze the effect of \(r \in \{10,20,30,40,50\}\) on Cora and Computers with \(s=3\) and \(l=4\). As shown in Figure~\ref{fig:sampling}, model performance quickly improves and then plateaus as \(r\) increases, while training time grows substantially. On both datasets, \(r=30\) achieves accuracy comparable to \(r=50\) with 2.2\(\times\) and 1.6\(\times\) training speedups, respectively. These results suggest that sampling a small set of structurally central nodes is sufficient for subgraph-level learning. We therefore adopt \(r=30\) in the main experiments. Results on additional datasets are provided in Appendix~\ref{app:ps}.

\section{Related Work}
Our work is closely related to self-supervised learning on hypergraphs and representation learning on text-attributed graphs (see Appendix~\ref{app:tag}).
\subsection{Self-supervised Learning on Hypergraphs}
SSL has emerged as a promising paradigm for hypergraph representation learning without expensive labels.
Most existing HSSL methods are based on contrastive learning.
HyperGCL~\cite{hypergcl2022} introduces a learnable augmentation strategy, while TriCL~\cite{tricl2023} and SE-HSSL~\cite{se-hssl2024} design hierarchical contrastive objectives across multiple granularities.
Beyond contrastive methods, generative SSL has also been explored, with HypeBoy~\cite{hypeboy2024} proposing a hyperedge-filling task. 
Other efforts explore alternative settings, such as feature-absent scenarios in VilLain~\cite{villain2024}.
While effective for modeling hypergraphs, these methods largely overlook the rich textual semantics in TAHGs, resulting in suboptimal performance.
HyperBERT~\cite{hyperbert2024} pioneers SSL on TAHGs by jointly optimizing text and hypergraph encoders through contrastive learning. However, its contrastive signals are limited to 1-hop hyperedge relations, restricting the modeling of long-range dependencies.
\section{Conclusion}
We propose HiTeC, an effective contrastive learning framework tailored for TAHGs.
HiTeC adopts a two-stage paradigm that first injects topological information into text representations and then performs semantic-aware hierarchical contrastive learning on TAHGs. Specifically,  two semantic-aware augmentation strategies are introduced to reduce perturbation noise, while an \(s\)-walk-based subgraph-level contrast captures long-range dependencies beyond node- and hyperedge-level scopes.
Extensive experiments on six real-world datasets and multiple downstream tasks demonstrate the effectiveness and robustness of HiTeC. 

\clearpage
\newpage
\section*{Limitations}
There are two limitations in our work:
(1) The construction of the structure-contextualized text view relies on a handcrafted template. Although the proposed template demonstrates strong empirical effectiveness across multiple datasets, its optimal form may vary across different domains and tasks. Future work may explore more adaptive text-level augmentation.
(2) The structure-aware pre-training stage introduces additional training overhead compared with conventional HCL methods. Nevertheless, HiTeC remains more efficient than prior TAHG methods such as HyperBERT due to its decoupled training paradigm (see Appendix~\ref{app:efficiency}). 
\bibliography{custom}

\clearpage
\newpage
\appendix
\section*{Appendix}
\section{Additional Template Details}\label{sec:template}
\begin{table*}[t]
\centering
\footnotesize
\renewcommand{\arraystretch}{0.9}
\setlength{\tabcolsep}{6pt}

\begin{tabular}{
    >{\centering\arraybackslash}m{0.10\textwidth}
  | m{0.15\textwidth}
  | >{\raggedright\arraybackslash}m{0.65\textwidth}
}
\toprule
\textbf{Component} & \textbf{Description} & \textbf{Template} \\
\midrule
\textbf{[DK]} & Domain Knowledge &
You are in an \opt{Academic / E-commerce / \dots} hypergraph.
Each node represents a \opt{Paper / Product / \dots},
and hyperedges capture \opt{Co-citation / Co-purchasing / \dots}
relationships among groups of nodes.
\\
\midrule
\textbf{[GT]} & Global Topology &
Your node degree is \opt{value}, indicating a
\opt{Sparse / Dense} connectivity pattern.
You participate in the following hyperedges:
\opt{Hyperedge IDs}.
 \\
\midrule
\textbf{$t$} & Raw Text &
Your title is \opt{Title}.
Your abstract or description is \opt{Abstract / Description}.
\\
\midrule
\textbf{[NC]} & Neighbor Context &
Below are brief descriptions of your neighboring nodes.
For neighbor 1, the title is: \opt{Neighbor 1 Title}
and the abstract or description is :\opt{Neighbor 1 Abstract / Description}.
For neighbor 2 \dots\\
\bottomrule
\end{tabular}

\caption{Structure-contextualized text view template. }
\label{tab:template}
\end{table*}
As shown in Table~\ref{tab:template}, the structure-contextualized view follows a fixed template that concatenates contextual components into a single sequence:
$ \tilde{t}_{1,i} =
[\texttt{[DK]} \,||\, \texttt{[GT]} \,||\, \texttt{[NC]} \,||\, t_i].
$
All components are deterministically constructed from the hypergraph structure and node attributes, without external knowledge.
The number of neighboring nodes in \texttt{[NC]} is limited to 10 to control sequence length and ensure stable training.
This design enriches structural and semantic context while preserving semantic consistency.

\section{Complexity Analysis}
\label{app:complexity}
We analyze the computational complexity of the forward phase of HiTeC during training.

\myparagraph{Text encoder pretraining.}
Let $l_p$ denote the number of layers in the PLM, $\bar L$ the average token length, and $d$ the hidden dimension.
The PLM encoding costs
$\mathcal{O}\!\left(l_p |\mathcal{V}| (\bar L d^2 + \bar L^2 d)\right)$.
The triplet margin objective costs $\mathcal{O}(|\mathcal{V}|\, d)$.
Therefore, the overall complexity is
$\mathcal{O}\!\left(l_p |\mathcal{V}| (\bar L d^2 + \bar L^2 d) + |\mathcal{V}| d\right)$.

\myparagraph{Hypergraph encoder pretraining.}
Let $l_h$ be the number of HNN layers and $\delta_e$ the average hyperedge size.
The hypergraph propagation costs
$\mathcal{O}\!\left(l_h (|\mathcal{E}| \delta_e d + |\mathcal{V}| d^2)\right)$.
For contrastive learning with $m$ sampled negatives (a constant),
the node-, hyperedge-, and subgraph-level objectives incur costs of
$\mathcal{O}(m |\mathcal{V}| d)$,
$\mathcal{O}(m |\mathcal{E}| d)$,
and $\mathcal{O}(m K d)$, respectively,
where $K$ denotes the number of sampled subgraphs.
Thus, the total complexity is $\mathcal{O}(l_h(|\mathcal{E}|\delta_e d+|\mathcal{V}| d^{2}) + m(|\mathcal{V}|+|\mathcal{E}|+K)d)$.

\section{Additional Experimental Setup}

\subsection{Datasets}\label{app:dataset}
Since no publicly available benchmark datasets are specifically designed for Text-Attributed Hypergraphs (TAHGs), we construct six datasets spanning two domains: academic co-citation networks and E-commerce platforms. Basic statistics are summarized in Table~\ref{dataset}.

\myparagraph{Co-citation hypergraphs.}
We construct two academic hypergraphs, Citeseer~\cite{citeseer} and Cora~\cite{cora}, from standard citation networks, where each node represents a scientific publication. Node textual attributes, including titles and abstracts, are adopted from prior works~\cite{tape2024,graphbridge2024}. To capture high-order relations, we define each hyperedge as the set of papers co-citing the same reference.

\myparagraph{E-commerce hypergraphs.}
We further construct four e-commerce hypergraphs from the Amazon product network~\cite{amazon}: Book-History, Ele-Photo, Ele-Computers, and Sports-Fitness. Each node represents a product, while edges reflect frequent co-purchase or co-browsing behavior. Node textual attributes are adopted from~\cite{cs-tag2023}. Specifically, Book-History uses book titles and descriptions; Ele-Computers and Ele-Photo use high-rated reviews and product summaries; Sports-Fitness only uses product titles. To induce high-order structures, we adopt a max-clique-based reconstruction strategy~\cite{maxclique1973,reconstruction2024}, where hyperedges are formed from maximal cliques in the original graph.

\subsection{Baselines}\label{app:baseline}
We compare our method against eight representative self-supervised learning (SSL) baselines, covering graph-based, hypergraph-based, and text-attributed hypergraph (TAHG)-based paradigms.

\myparagraph{Graph-based SSL.} In this work, we adopt three widely used graph-based SSL frameworks: two contrastive learning methods, GraphCL~\cite{graphcl2020} and BGRL~\cite{bgrl2022}, and one generative method, GraphMAE~\cite{graphmae2022}. To adapt these methods to hypergraphs, we apply clique expansion following the protocol in~\cite{tricl2023}. For textual feature extraction, we employ shallow encoders (SE) implemented via Skip-Gram~\cite{skip2013}, as well as GIANT~\cite{giant2022}, a graph-agnostic text encoder designed for raw textual attributes.

\myparagraph{Hypergraph-based SSL.}
We consider four state-of-the-art self-supervised learning methods specifically designed for hypergraphs.
These include two contrastive approaches, TriCL~\cite{tricl2023} and SE-HSSL~\cite{se-hssl2024};
one generative framework, HypeBoy~\cite{hypeboy2024};
and VilLain~\cite{villain2024}, which is tailored for feature-absent hypergraph settings.
For a fair comparison, all methods adopt the same shallow encoder (SE).
In addition, we further evaluate these baselines by replacing the SE with two pre-trained language models,
BERT-Base~\cite{bert2019} and RoBERTa-Base~\cite{roberta2019}.

\myparagraph{TAHG-based SSL.}
We further include HyperBERT~\cite{hyperbert2024}, to the best of our knowledge, the only existing SSL method specifically designed for text-attributed hypergraphs.
HyperBERT jointly models textual semantics and hypergraph structure using a Transformer-based architecture, in which hypergraph-aware layers are introduced to integrate structural signals with text representations.
However, this joint modeling paradigm incurs substantial computational and memory overhead, limiting its scalability to large hypergraphs, as validated in our experiments.

\subsection{Evaluation Protocol}\label{app:ep}
We evaluate our method on two standard hypergraph learning tasks: node classification and hyperedge prediction. After pretraining with HiTeC, we adopt a linear evaluation protocol~\cite{tricl2023,se-hssl2024,hypeboy2024}, where the pretrained node embeddings are frozen and used as input features for downstream classifiers.

\myparagraph{Node classification.}
Following~\cite{tricl2023,se-hssl2024}, we randomly split each dataset into 10\%/10\%/80\% for training, validation, and testing. Then, a simple linear classifier is trained on top of the fixed node embeddings using \(\ell_2\)-regularized logistic regression. We report the average test accuracy and standard deviation over 20 random splits, each with 5 different initializations.

\myparagraph{Hyperedge prediction.}
Following~\cite{hypeboy2024}, we need to generate negative hyperedge samples to evaluate a model on the hyperedge prediction task. In our experiments, we adopt Clique Negative Sampling (CNS)~\cite{hygen2025}, a widely used strategy for constructing negative hyperedge samples. Specifically, for a given hyperedge \(e\), we replace a node \(u \in e\) with a node \(v \notin e\) such that \(v\) is connected to all other nodes in \(e\), i.e., \( (e \setminus \{u\}) \cup \{v\} \). Then, a two-layer MLP is trained on the fixed node embeddings using the 60\%/20\%/20\% data split. Same as the node classification task, we report the average test accuracy and standard deviation over 20 random splits, each with 5 initializations.

\subsection{Implementation Details}\label{app:imp}
All experiments are conducted on a single NVIDIA A800 PCIe GPU with 80GB of memory.
For all baselines, we adopt their official implementations with default settings. 
For HiTeC, we adopt BERT-Base~\cite{bert2019} as the default text encoder and apply LoRA~\cite{lora2022} to the query and value matrices of the self-attention layers for parameter-efficient fine-tuning, while keeping the PLM backbone frozen. Unless otherwise specified, the LoRA configuration is set to \(r=16\), \(\alpha=32\), and dropout \(=0.05\). We additionally evaluate other text encoder backbones, including DistilBERT~\cite{distilbert2019}, RoBERTa-Base~\cite{roberta2019}, and DeBERTa-Base~\cite{deberta2020}.
The text encoder is trained for 5 epochs on small datasets (Citeseer, Cora), 4 epochs on medium datasets (History, Photo), and 3 epochs on large datasets (Computers, Fitness), using a learning rate of 2e-5. 
The hypergraph encoder is implemented as a single-layer module. We tune the contrastive loss weights \(\lambda_e\) and \(\lambda_s\) over \( \{1,2,3,4 \}\). The overlap threshold \( s \) and walk length $l$ are tuned from \(\{1, 2, 3, 4, 5, 10\}\), the subgraph sampling ratio \( r \) in \(\{10, 20, 30, 40, 50\}\).
To ensure a fair comparison, we tune the learning rate from \(\{0.01, 0.001, 1e\text{-}4, 5e\text{-}4\}\) and weight decay from \(\{0, 1e\text{-}5\}\) for all methods.

\section{Additional Experiment Results}
\begin{table*}
\centering
\footnotesize 
\renewcommand{\arraystretch}{0.9} 
\setlength{\tabcolsep}{4pt}
\begin{tabular}{llcccccc}
\toprule
\textbf{} & \textbf{Method} & \textbf{Citeseer} & \textbf{Cora} & \textbf{History} & \textbf{Photo} & \textbf{Computers} & \textbf{Fitness} \\
\midrule
\midrule
\multirow{6}{*}{\rotatebox{90}{\textbf{Graph SSL}}}
&GraphCL + SE         & 69.48 ± 2.07 & 59.05 ± 2.16 & 65.65 ± 1.06 & 67.48 ± 1.18 & 68.66 ± 1.33 & 61.22 ± 1.34\\
&GraphCL + GIANT      & 72.70 ± 2.06 & 55.48 ± 1.96 & 63.02 ± 1.61 & 68.78 ± 1.16 & 70.87 ± 0.87 & 63.02 ± 0.01 \\
&BGRL+SE            & 70.08 ± 2.38 & 62.45 ± 2.48 & 64.47 ± 1.18 & 64.76 ± 1.23 & 64.66 ± 1.40 & 60.05 ± 1.05 \\
&BGRL + GIANT         & 72.61 ± 1.99 & 66.22 ± 2.38 & 65.13 ± 2.09 & 65.65 ± 1.10 & 67.67 ± 0.99 & 60.92 ± 0.34 \\
&GraphMAE + SE        & 71.71 ± 1.78 & 64.85 ± 1.66 & 59.81 ± 0.27 & 66.33 ± 1.58 & 64.83 ± 1.72 & 55.79 ± 0.18  \\
&GraphMAE + GIANT     & 72.25 ± 1.62 & 66.30 ± 2.19 & 64.19 ± 0.23 & 64.83 ± 0.20 & 66.10 ± 1.20 & 59.21 ± 0.88 \\
\midrule
\multirow{11}{*}{\rotatebox{90}{\textbf{HSSL}}}
&TriCL + SE      & 68.06 ± 1.75 & 62.43 ± 2.41 & 67.18 ± 0.24 & 69.21 ± 0.19 & 70.63 ± 0.14  &  63.87 ± 0.30 \\
&TriCL + BERT    & 71.36 ± 1.78 & 64.02 ± 2.33 & \underline{70.18 ± 0.31} & \underline{71.71 ± 0.21} & \underline{73.20 ± 0.17}  & 65.49 ± 0.40 \\
&TriCL + RoBERTa & 72.38 ± 1.37 & 66.49 ± 1.93 & 68.34 ± 0.58  & 70.35 ± 0.28  & 72.20 ± 0.27  & 63.49 ± 0.17 \\
&SE-HSSL + SE      & 64.80 ± 2.21 & 65.57 ± 1.88 & 63.49 ± 1.98 & 65.42 ± 0.74 & 66.44 ± 1.09 & 62.52 ± 0.18 \\
&SE-HSSL + BERT    & 72.29 ± 1.74 & 65.66 ± 2.25 & 67.93 ± 0.44 & 71.27 ± 0.18 & 70.68 ± 0.91 & \underline{66.00 ± 0.69} \\
&SE-HSSL + RoBERTa & 73.09 ± 1.56 & 65.86 ± 1.94 & 66.95 ± 0.53 & 69.65 ± 0.20 & 71.19 ± 0.16 & 65.72 ± 0.70 \\
&HypeBoy + SE      & 71.93 ± 2.04 & 65.17 ± 1.66 & OOM & OOM & OOM & OOM \\
&HypeBoy + BERT    & \underline{74.56 ± 2.10} & \textbf{67.38 ± 1.89} & OOM & OOM & OOM & OOM \\
&HypeBoy + RoBERTa & 73.47 ± 2.10 & \underline{66.51 ± 1.87} & OOM & OOM & OOM & OOM \\
&VilLain (w/o feat.) & 74.25 ± 1.90 & 63.25 ± 2.57 & 63.18 ± 0.35 & 65.02 ± 0.23 & 65.85 ± 0.28 &  62.92 ± 0.26 \\
&HyperBERT       & 71.84 ± 2.03 & 64.48 ± 2.10 & 69.52 ± 0.20 & 69.34 ± 0.18 & 70.40 ± 0.15 & OOM \\
\cmidrule(lr){2-8}
&HiTeC & \textbf{75.22 ± 1.87} & 65.91 ± 2.32 & \textbf{71.83 ± 0.21} & \textbf{72.56 ± 0.17} & \textbf{73.91 ± 0.14} & \textbf{67.32 ± 0.09}\\
\bottomrule
\end{tabular}
\caption{Evaluation results on hyperedge prediction task(\% mean \(\pm\) std). The \textbf{best} and \underline{second-best} results are highlighted in \textbf{bold} and \underline{underline}, respectively. ``w/o feat.'' denotes methods without node features. OOM indicates that the method runs out of memory on a single 80GB GPU.}
\label{tab:hyperedge}
\end{table*}
\subsection{Hyperedge Prediction Evaluation}\label{app:hp}
We further evaluate all models on the hyperedge prediction task, with results presented in Table~\ref{tab:hyperedge}. The key observations are as follows: (1) Graph-based SSL methods remain less effective. Their performance is still hindered by the loss of high-order structure due to clique expansion. Although GIANT achieves slight improvements, probably due to its pseudo-labeling strategy for edge-level supervision, it still underperforms compared to HSSL methods enhanced with pre-trained language models.
(2) HiTeC achieves the best performance on 5 out of 6 datasets, demonstrating strong generalizability between tasks. TriCL and SE-HSSL underperform due to their limited capacity to capture long-range dependencies. HypeBoy exhibits competitive results on small-scale hypergraphs, benefiting from its hyperedge-filling strategy, but fails to scale to larger datasets, indicating poor scalability.
Overall, HiTeC exceeds the second-best method by 1.65\% and 1.15\% on History and Fitness, respectively, demonstrating its robustness and scalability across diverse hypergraph scenarios.

\subsection{Impact of Text Encoder Backbones}\label{app:backbones}
We evaluate different text encoder backbones, including DistilBERT~\cite{distilbert2019}, BERT-Base~\cite{bert2019}, RoBERTa-Base~\cite{roberta2019}, and DeBERTa-Base~\cite{deberta2020}, as shown in Table~\ref{tab:backbone}. Among them, BERT-Base achieves the best overall performance, followed by DistilBERT. The results indicate that backbone selection can noticeably affect performance across datasets. Nevertheless, our framework remains compatible with diverse text encoders. Based on the overall results, we adopt BERT as the default backbone in the main experiments.
\begin{table}
  \centering
  \small
   \begin{adjustbox}{width=\linewidth}
    \begin{tabular}{cccccc}
    \toprule
    \textbf{Dataset} & \textbf{DistilBERT} & \textbf{BERT} & \textbf{RoBERTa}  & \textbf{DeBERTa}\\
    \midrule
    Citeseer    & \textbf{67.54 ± 1.49}  & \underline{66.38 ± 1.73} & 65.09 ± 1.26 & 64.83 ± 4.65\\
    Cora        & \textbf{74.55 ± 0.02} & \underline{74.07 ± 0.88} & 72.32 ± 0.91 &  {70.41 ± 2.24} \\
    History     & \underline{78.05 ± 0.18}  & \textbf{79.81 ± 0.16} & 74.77 ± 0.24 & {76.20 ± 0.12} \\
    Photo       & \underline{56.90 ± 0.31} & \textbf{59.64 ± 0.16}  & 54.15 ± 0.40 & {52.54 ± 0.08}  \\
    Computers   & 50.64 ± 0.25 & \textbf{55.81 ± 0.12} & \underline{52.36 ± 0.21}  & {51.48 ± 0.56} \\
    Fitness     & 62.97 ± 0.08 & \textbf{67.18 ± 0.19} & 64.98 ± 0.78  & \underline{65.29 ± 0.14}\\
    \bottomrule
    \end{tabular}
    \end{adjustbox}
  \caption{\label{tab:backbone}
    Different text encoder backbones. 
  }
\end{table}

\subsection{ Parameter Sensitivity Analysis of \(s\)-walk length \(l\)}\label{app:length}
\begin{figure}[t]
\centering
\includegraphics[width=1\columnwidth]{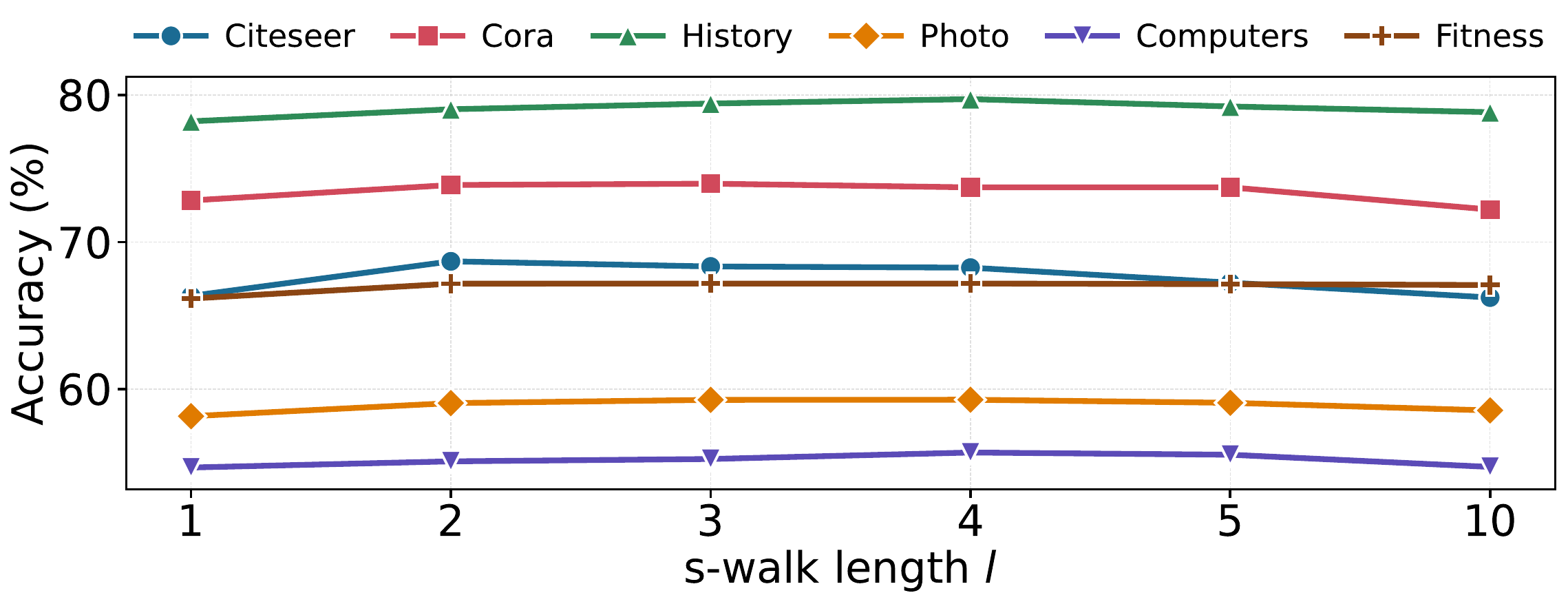} %
\caption{Sensitivity analysis of walk length $l$. }
\vspace{-1em}
\label{fig:s-walk-length}
\end{figure}
To investigate the effect of walk length on HiTeC, we report the results under different \(l \in \{1,2,3,4,5,10\}\) while fixing \(s=3\). 
As shown in Figure~\ref{fig:s-walk-length}, model performance generally first improves and then degrades as \(l\) increases. This trend suggests that moderate walk lengths help capture richer high-order structural dependencies by exploring broader hypergraph contexts. In contrast, short walks (e.g., \(l=1\)) provide limited structural context, while excessively long walks (e.g., \(l=10\)) may introduce noisy or weakly related hyperedges, both leading to suboptimal performance. Overall, \(l=4\) achieves a favorable balance between structural exploration and semantic consistency.

\subsection{ Parameter Sensitivity Analysis of Subgraph Sampling Ratio \(r\)}\label{app:ps}
\begin{figure}[ht]
\centering
\includegraphics[width=1\columnwidth]{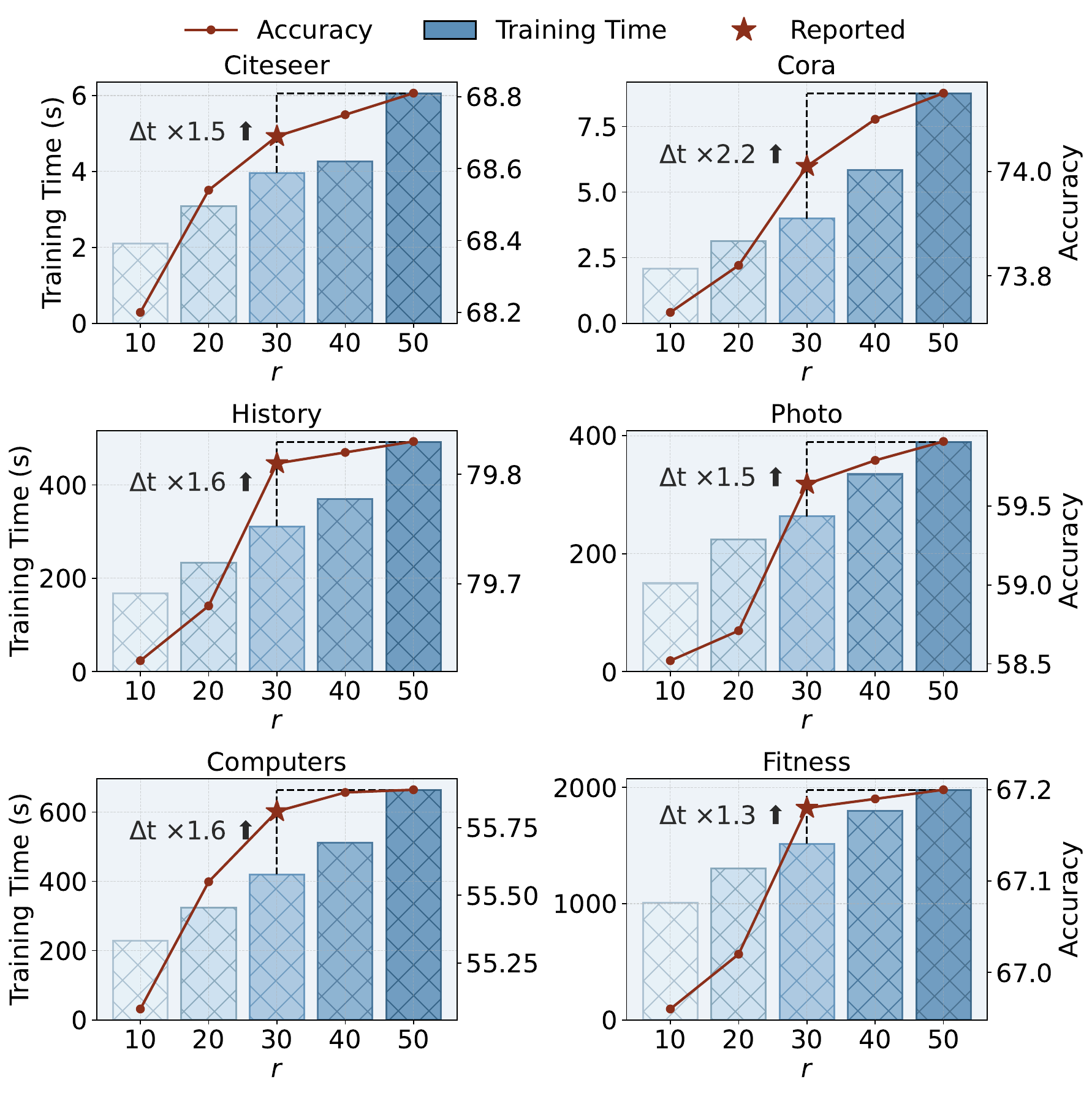} 
\caption{
Training time and accuracy under different sampling ratios $r$.
Bar plots report training time (in seconds), while line plots show the corresponding accuracy.
We highlight the relative speedup $\Delta t$ of $r=30$ over $r=50$. $s$ and $l$ are fixed to 3 and 4, respectively. All main results are selected with $r=30$.
}
\label{sub_efficiency}
\end{figure}

We further perform a sensitivity analysis by varying the subgraph sampling ratio $r \in \{10, 20, 30, 40, 50\}$ with $s$ fixed as 3 and $l$ fixed as 4. As shown in Figure~\ref{sub_efficiency}, model performance generally increases rapidly at first and then plateaus, while training time grows sharply with larger $r$. 
Notably, on most datasets, setting $r=30$ achieves comparable accuracy to $r=50$, while significantly reducing training time. For example, on Cora and Computers, $r=30$ leads to 2.2$\times$ and 1.6$\times$ speedups, respectively, with only marginal drops in accuracy. These results indicate that selecting a small number of structurally central nodes based on degree ranking enables efficient subgraph-level representation learning without sacrificing performance. In our main experiments, we report the results with \(r=30\), which achieves a trade-off between efficiency and performance.

\subsection{Efficiency Evaluation}\label{app:efficiency}
We evaluate training time to assess the efficiency of HiTeC. 
Specifically, we first compare the text encoder pre-training time between HiTeC and HyperBERT, and then compare the hypergraph encoder pre-training time of HiTeC with two representative HCL methods (TriCL and SE-HSSL).

\myparagraph{Text encoder pre-training time evaluation.}
Table~\ref{text-pre-training-time} compares the text encoder pre-training time between HyperBERT and HiTeC. For HyperBERT, we follow the official implementation and report the converged training time under its recommended training settings. For HiTeC, we adopt the settings described in the Appendix~\ref{app:imp}. 
From the results, HiTeC consistently achieves substantially lower pre-training time across all datasets. 
In particular, HiTeC achieves nearly 2$\times$ faster pre-training on large datasets such as Computers (56h 04m \(\rightarrow\) 28h 44m), while still achieving superior downstream performance.
This efficiency gain mainly comes from the decoupled pre-training paradigm of HiTeC, whereas HyperBERT relies on a joint training scheme for text and hypergraph modeling.
\begin{table}
  \centering
   \begin{adjustbox}{width=\linewidth}
    \begin{tabular}{ccccccc}
    \toprule
    \textbf{Method} & \textbf{Citeseer} & \textbf{Cora} & \textbf{History} & \textbf{Photo} & \textbf{Computers} \\
    \midrule
    HyperBERT  & 4h 30m & 5h 22m  & 29h 42m  & 33h 40m &  56h 04m   \\
    HiTeC      & 2h 22m & 2h 56m  & 18h 31m  & 19h 55m &  28h 44m  \\
    \bottomrule
    \end{tabular}
    \end{adjustbox}
  \caption{\label{text-pre-training-time}
   Comparison of text encoder pretraining time between HyperBERT and HiTeC.
  }
  \vspace{-1em}
\end{table}

\begin{figure}[t]
\centering
\includegraphics[width=1\columnwidth]{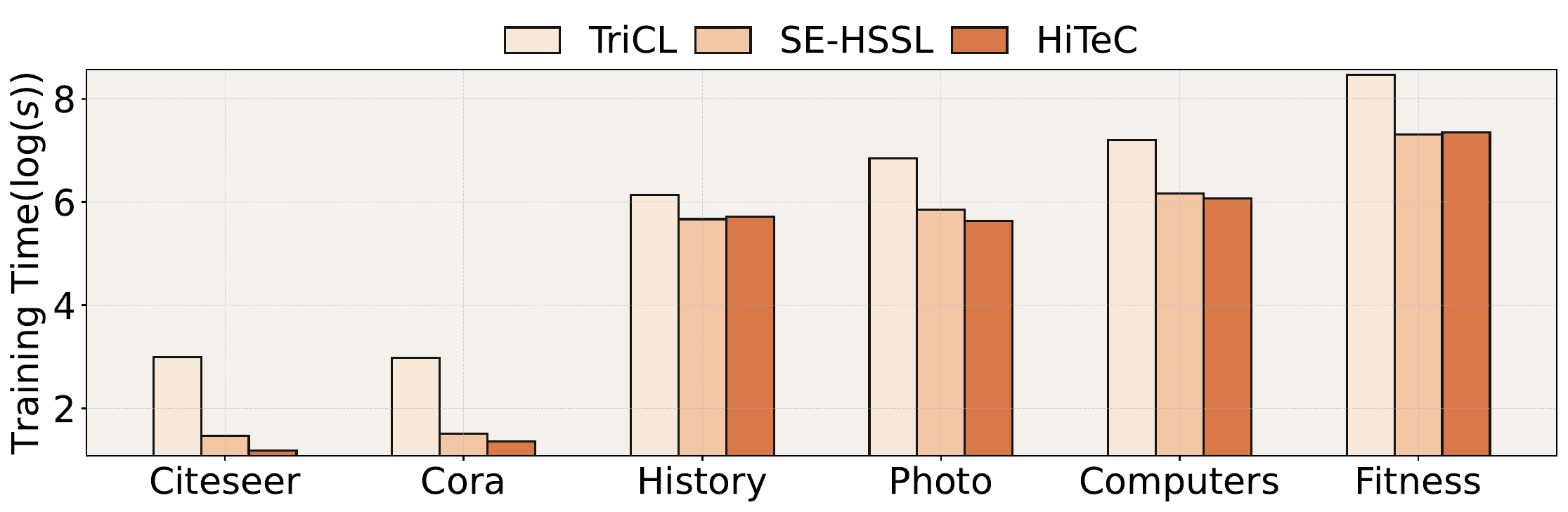} %
\caption{Comparison of hypergraph encoder pretraining time among HiTeC, SE-HSSL, and TriCL. log(·) denotes the natural logarithm.}
\vspace{-1em}
\label{hgnn-training-time}
\end{figure}

\myparagraph{Hypergraph encoder pretraining time evaluation.}
We further compare the hypergraph encoder pre-training time of HiTeC with two representative hypergraph-based HCL methods.
As shown in Figure~\ref{hgnn-training-time}, training time is plotted on a logarithmic scale for clearer visualization, where HiTeC achieves performance comparable to SE-HSSL while being at least $3\times$ faster than TriCL on large-scale datasets.
This efficiency gain mainly stems from two factors:
(1) The subgraph-level contrastive objective operates on a small subset of sampled subgraphs (see Appendix~\ref{app:ps}); and
(2) The preprocessing steps are executed once and reused across multiple training runs.
Overall, HiTeC strikes a favorable trade-off between training efficiency and accuracy.

\section{Additional Related Work}
\subsection{Representation Learning on TAGs}\label{app:tag}
Since hypergraphs can be transformed into graphs via clique expansion, our work is closely related to representation learning on text-attributed graphs (TAGs). 
Recent TAG methods leverage pre-trained language models to generate numerical features, typically in a cascade~\cite{tape2024,graphbridge2024} or nested~\cite{graphformers2021,glem2023} manner.
While most existing TAG methods are supervised, emerging efforts explore self-supervised learning (SSL) to reduce annotation costs. GIANT~\cite{giant2022} makes an early attempt by aligning textual and structural features via an extreme multi-label classification objective. More recently, GAugLLM~\cite{gaugllm2024} proposes an augmentation framework that utilizes large language models (LLMs) to perturb both node features and graph structure, while HASH-CODE~\cite{hash-code2024} formulates a multi-objective contrastive learning strategy tailored for TAGs.
However, hypergraphs fundamentally differ from ordinary graphs in their high-order structural modeling. Converting hypergraphs into ordinary graphs via clique expansion~\cite{maxclique1973} inevitably incurs substantial structural information loss~\cite{tricl2023,se-hssl2024}, limiting their effectiveness on TAHGs and motivating the need for dedicated HSSL approaches.

\end{document}